\def\year{2017}\relax
\documentclass[letterpaper]{article}
\usepackage{aaai17}
\usepackage{times}
\usepackage{helvet}
\usepackage{courier}

\usepackage[utf8]{inputenc} % allow utf-8 input
\usepackage[T1]{fontenc}    % use 8-bit T1 fonts
\usepackage{url}            % simple URL typesetting
\usepackage{amsfonts}       % blackboard math symbols
\usepackage{microtype}      % microtypography
\usepackage[pdftex]{graphicx}
\usepackage{algorithm}
\usepackage{algorithmic}
\usepackage{amsmath}
\usepackage{amssymb}
\usepackage{mathrsfs}
\usepackage{array}
\usepackage{bm}
\usepackage{multirow}
\usepackage{natbib}
\usepackage{subfig}

\def\E{{\rm E}}
\def\P{P_{\rm data}}
\def\KL{{\rm KL}}
\def\S{{\bf S}}

\begin{document}
\title{Alternating Back-Propagation for Generator Network}
\author{Tian Han $^\dagger$, Yang Lu $^\dagger$, Song-Chun Zhu, Ying Nian Wu\\
Department of Statistics, University of California, Los Angeles, USA}
\maketitle
\begin{abstract}
This paper proposes an alternating back-propagation algorithm for learning the generator network model. The model is a non-linear generalization of  factor analysis. In this model, the mapping from the continuous latent factors to the observed signal is parametrized by a convolutional neural network. The alternating back-propagation algorithm iterates the following two steps: (1) {\em Inferential back-propagation}, which infers the latent factors by Langevin dynamics or gradient descent. (2) {\em Learning back-propagation}, which updates the parameters given the inferred latent factors by gradient descent. The gradient computations in both steps are powered by back-propagation, and they share most of their code in common. We show that the alternating back-propagation algorithm can learn realistic generator models of natural images, video sequences, and sounds. Moreover, it can also be used to learn from incomplete or indirect training data. 
\end{abstract}
\let\thefootnote\relax\footnotetext{$^\dagger$ Equal contributions.}

\section{Introduction}

This paper studies the fundamental problem of learning and inference in the generator network \citep{goodfellow2014generative}, which is a generative model that has become popular recently. Specifically, we propose an alternating back-propagation algorithm for learning and inference in this model.

\subsection{Non-linear factor analysis}

The generator network is a non-linear generalization of factor analysis. Factor analysis is a prototype model in unsupervised learning of distributed representations. There are two directions one can pursue in order to generalize the factor analysis model. One direction is to generalize the prior model or the prior assumption about the latent factors. This led to methods such as independent component analysis \citep{hyvarinen2004independent}, sparse coding \citep{olshausen1997sparse}, non-negative matrix factorization \citep{lee2001algorithms}, matrix factorization and completion for recommender systems \citep{koren2009matrix}, etc. 

The other direction to generalize the factor analysis model is to generalize the mapping from the continuous latent factors to the observed signal.  The generator network is an example in this direction. It generalizes the linear mapping in factor analysis to a non-linear mapping that is defined by a convolutional neural network (ConvNet or CNN) \citep{lecun1998gradient,krizhevsky2012imagenet,Alexey2015}. It has been shown recently that the generator network is capable of generating realistic images \citep{denton2015deep,radford2015unsupervised}. 

The generator network is a fundamental representation of knowledge, and it has the following properties: (1) {\em Analysis}: The model disentangles the variations in the observed signals  into independent variations of latent factors. (2) {\em Synthesis}: The model can synthesize new signals by sampling the factors from the known prior distribution and transforming the factors into the signal.  (3) {\em Embedding}: The model embeds the high-dimensional non-Euclidean manifold formed by the observed signals into the low-dimensional Euclidean space of the latent factors, so that linear interpolation  in the low-dimensional factor space results in non-linear interpolation in the data space.

\subsection{Alternating back-propagation} 

The factor analysis model can be learned by the Rubin-Thayer EM algorithm  \citep{rubin1982algorithms,dempster1977maximum}, where both the E-step and the M-step are based on multivariate linear regression.  Inspired by this algorithm, we propose an alternating back-propagation algorithm for learning the generator network that iterates the following two-steps: 

(1) {\em  Inferential back-propagation}: For each training example,  infer the continuous latent factors by Langevin dynamics or gradient descent. 

(2) {\em Learning back-propagation}: Update the parameters given the inferred latent factors by gradient descent. 

The Langevin dynamics \citep{neal2011mcmc} is a stochastic sampling counterpart of gradient descent. The gradient computations in both steps are powered by back-propagation. Because of the ConvNet structure, the gradient computation in step (1) is actually a by-product of the gradient computation in step (2) in terms of coding. 

Given the factors, the learning of the ConvNet is a supervised learning problem \citep{Alexey2015} that can be accomplished by the learning back-propagation. With factors unknown, the learning becomes an unsupervised problem, which can be solved by adding the inferential back-propagation as an inner loop of the learning process.  We shall show that the alternating back-propagation algorithm can learn realistic generator models of natural images, video sequences, and sounds. 

The alternating back-propagation algorithm follows the tradition of alternating operations in unsupervised learning, such as alternating linear regression in the EM algorithm for factor analysis,   alternating least squares algorithm for matrix factorization \citep{koren2009matrix,kim2008nonnegative}, and  alternating gradient descent algorithm for sparse coding \citep{olshausen1997sparse}. All these unsupervised learning algorithms alternate an inference step and a learning step, as is the case with alternating back-propagation. 

\subsection{Explaining-away inference}

The inferential back-propagation solves an inverse problem by an explaining-away process, where the latent factors compete with each other to explain each training example.  The following are the advantages of the explaining-away inference of the latent factors: 

(1) The latent factors may follow sophisticated prior models. For instance, in textured motions \citep{Wang03modelingtextured} or dynamic textures \citep{dorettoCWS03}, the latent factors may follow a dynamic  model such as vector auto-regression.  By inferring the latent factors that explain the observed examples, we can learn the prior model.

(2) The observed data may be {incomplete} or {indirect}.  For instance, the training images may contain occluded objects. In this case, the latent factors can still be obtained by explaining the incomplete or indirect observations, and the model can still  be learned as before.

\subsection{Learning from incomplete or indirect data}

We venture to propose that a main advantage of a generative model is to learn from incomplete or indirect data, which are not uncommon in practice.  The generative model can then be evaluated based on how well  it recovers the unobserved original data, while still learning a model that can generate new data. Learning the generator network from incomplete data can be considered a non-linear generalization of matrix completion. 

We also propose to evaluate the learned generator network by the reconstruction error on the testing data. 

\subsection{Contribution and related work}

The main contribution of this paper is to propose the alternating back-propagation algorithm for training the generator network. Another contribution is to evaluate the generative models by learning from incomplete or indirect training data.  

Existing training methods for the generator network avoid explain-away inference of latent factors. Two methods have recently been devised to accomplish this. Both methods involve an assisting network with a separate set of parameters in addition to the original network that generates the signals. One method is variational auto-encoder (VAE) \citep{KingmaCoRR13,RezendeICML2014,MnihGregor2014}, where the assisting  network is an inferential or recognition network that seeks to approximate the posterior distribution of the latent factors. 
%This method can be traced back to the wake-sleep algorithm \citep{hinton1995wake}.  
The other method is the generative adversarial network (GAN) \citep{goodfellow2014generative,denton2015deep,radford2015unsupervised}, where the assisting network is a discriminator network that plays an adversarial role against the generator network. 

Unlike alternating back-propagation, VAE does not perform explicit explain-away inference, while GAN avoids inferring the latent factors altogether. In comparison, the alternating back-propagation algorithm is simpler and more basic, without  resorting to an extra network. While it is difficult to compare these methods directly, we illustrate the strength of  alternating back-propagation by learning from incomplete and indirect data, where we only need to explain whatever data we are given. This may prove difficult or less convenient for VAE and GAN. 

Meanwhile,  alternating back-propagation  is complementary to VAE and GAN training. It may use VAE to initialize the inferential back-propagation, and as a result,  may improve the inference in VAE. The inferential back-propagation may help infer the latent factors of the observed examples for GAN, thus providing a method to test if GAN can explain the entire training set. 

The generator network is based on a top-down ConvNet. One can also obtain a probabilistic model based on a bottom-up ConvNet that defines descriptive features \citep{XieLuICML,LuZhuWu2016}. 

\section{Factor analysis with ConvNet}

\subsection{Factor analysis and beyond} 

Let $Y$ be a $D$-dimensional observed data vector, such as an image. Let $Z$ be the $d$-dimensional vector of continuous latent factors, $Z  = (z_k, k = 1, ..., d)$. The traditional factor analysis model is $Y = W Z + \epsilon$, where $W$ is $D \times d$ matrix, and $\epsilon$ is a $D$-dimensional error vector or the observational noise. We assume that $Z \sim {\rm N}(0, I_d)$, where $I_d$ stands for the $d$-dimensional identity matrix. We also assume that $\epsilon \sim {\rm N}(0, \sigma^2 I_D)$, i.e., the observational errors are Gaussian white noises. There are three perspectives to view $W$. (1) {\em Basis vectors}. Write $W = (W_1, ..., W_d)$, where each $W_k$ is a $D$-dimensional column vector. Then $Y = \sum_{k=1}^{d} z_k W_k + \epsilon$, i.e., $W_k$ are the basis vectors and $z_k$ are the coefficients. (2) {\em Loading matrix}. Write $W = (w_1, ..., w_D)^\top$, where $w_j^\top$ is the $j$-th row of $W$. Then $y_j = \langle w_j, Z\rangle + \epsilon_j$, where $y_j$ and $\epsilon_j$ are the $j$-th components of $Y$ and $\epsilon$ respectively. Each $y_j$ is a loading of the $d$ factors where $w_j$ is a vector of loading weights,  indicating which factors are important for determining $y_j$. $W$ is called the loading matrix.  (3) {\em Matrix factorization}. Suppose we observe ${\bf Y} = (Y_1, ..., Y_n)$, whose factors are ${\bf Z} = (Z_1, ..., Z_n)$, then ${\bf Y} \approx W {\bf Z}$. 

The factor analysis model can be learned by the Rubin-Thayer EM algorithm, which involves alternating regressions of $Z$ on $Y$ in the E-step and of $Y$ on $Z$ in the M-step, with both steps powered by the sweep operator \citep{rubin1982algorithms,liu1998parameter}. 

The factor analysis model is the prototype of many subsequent models that generalize the prior model of $Z$.  (1) {\em Independent component analysis} \citep{hyvarinen2004independent}, $d = D$, $\epsilon = 0$, and $z_k$ are assumed to follow independent heavy tailed distributions. (2) {\em Sparse coding} \citep{olshausen1997sparse}, $d > D$, and $Z$ is assumed to be a redundant but sparse vector, i.e., only a small number of $z_k$ are non-zero  or significantly different from zero. (3) {\em Non-negative matrix factorization} \citep{lee2001algorithms}, it is assumed that $z_k \geq 0$. (4) {\em Recommender system} \citep{koren2009matrix}, $Z$ is a vector of a customer's desires in different aspects, and $w_j$ is a vector of product $j$'s desirabilities in these aspects. 

\subsection{ConvNet mapping} 

In addition to generalizing the prior model of the latent factors $Z$, we can also generalize the mapping from $Z$ to $Y$. In this paper, we consider the generator network model  \citep{goodfellow2014generative}  that retains the assumptions that $d < D$, $Z \sim {\rm N}(0, I_d)$, and $\epsilon \sim {\rm N}(0, \sigma^2 I_D)$ as in traditional factor analysis, but generalizes the linear mapping $W Z$ to a non-linear mapping $f(Z; W)$, where $f$ is a ConvNet, and $W$ collects all the connection weights and bias terms of the ConvNet. Then the model becomes
\begin{eqnarray}
&&Y = f(Z; W) + \epsilon, \nonumber \\
&&Z \sim {\rm N}(0,  I_d), \; \epsilon \sim {\rm N}(0, \sigma^2 I_D), \; d < D.
\label{eq:ConvNetFA}
\end{eqnarray}
The reconstruction error is  $ || Y - f(Z; W) ||^2$. We may assume more sophisticated models for $\epsilon$, such as colored noise or non-Gaussian texture. If $Y$ is binary, we can  emit $Y$ by a probability map $P = 1/[1+\exp(-f(Z; W))]$, where the sigmoid transformation and Bernoulli sampling are carried out pixel-wise. If $Y$ is multi-level, we may assume multinomial logistic emission model or some ordinal emission model. 

Although $f(Z; W)$ can be any non-linear mapping, the ConvNet parameterization of $f(Z; W)$ makes it particularly close to the original factor analysis. Specifically, we can write the top-down ConvNet as follows:
\begin{eqnarray}
Z^{(l-1)} = f_l(W_l Z^{(l)} + b_l), \label{eq:Conv2}
\end{eqnarray}
where $f_l$ is element-wise non-linearity at layer $l$, $W_l$ is the matrix of connection weights, $b_l$ is the vector of bias terms at layer $l $, and $W = (W_l, b_l, l = 1, ..., L)$.  $Z^{(0)} = f(Z; W)$, and $Z^{(L)} = Z$. The top-down ConvNet (\ref{eq:Conv2}) can be considered a recursion of the original factor analysis model, where the factors at the  layer $l-1$ are obtained by the linear superposition of the basis vectors or basis functions that are column vectors of $W_l$, with the factors at the layer $l$ serving as the coefficients of the linear superposition. 
In the case of ConvNet, the basis functions are shift-invariant versions of one another, like wavelets. See Appendix for an in-depth understanding of the model. 

\section{Alternating  back-propagation } 

If we observe a training set of data vectors $\{Y_i, i =  1, ..., n\}$, then each $Y_i$ has a corresponding $Z_i$, but all the $Y_i$ share the same ConvNet $W$.  Intuitively, we should infer $\{Z_i\}$ and learn $W$ to minimize the reconstruction error $\sum_{i=1}^{n}  || Y_i - f(Z_i; W) ||^2$ plus a regularization term that corresponds to the prior on $Z$.

More formally, the model can be written as $Z \sim p(Z)$ and $[Y|Z, W] \sim p(Y|Z, W)$. Adopting the language of the EM algorithm \citep{dempster1977maximum}, the complete-data model is given by \begin{eqnarray} 
&& \log p(Y, Z; W) = \log \left[p(Z) p(Y|Z, W) \right]\nonumber\\
&& = - \frac{1}{2\sigma^2} \|Y - f(Z; W)\|^2 - \frac{1}{2} \|Z\|^2 + {\rm const}.
\end{eqnarray}
The observed-data model is  obtained by integrating out $Z$: $p(Y; W) = \int p(Z) p(Y|Z, W) dZ$. The posterior distribution of $Z$ is given by $p(Z|Y, W) = p(Y, Z; W)/p(Y; W)  \propto p(Z) p(Y|Z, W)$ as a function of $Z$. 

For the training data $\{Y_i\}$, the complete-data log-likelihood is 
$L(W, \{Z_i\}) =  \sum_{i=1}^{n} \log p(Y_i, Z_i; W)$, where we assume $\sigma^2$ is given. Learning and inference can be accomplished by maximizing the complete-data log-likelihood, which can be obtained by the alternating gradient descent algorithm that iterates the following two steps: (1) Inference step: update $Z_i$ by running $l$ steps of gradient descent. (2) Learning step: update $W$ by one step of gradient descent. 

A more rigorous method is to maximize the observed-data log-likelihood, which is 
$ L(W) =  \sum_{i=1}^{n} \log p(Y_i; W)   = \sum_{i=1}^{n} \log \int p(Y_i, Z_i; W) dZ_i$.  The observed-data log-likelihood takes into account the  uncertainties in inferring $Z_i$. See Appendix for an in-depth understanding. 

The gradient of $L(W)$ can be calculated according to the following well-known fact that underlies the EM algorithm:
\begin{eqnarray} 
&&\frac{\partial}{\partial W} \log p(Y; W) =  \frac{1}{P(Y; W)}  \frac{\partial}{\partial W}  \int p(Y, Z; W) dZ \nonumber \\
%&&= \frac{1}{P(Y; W)} \int \left[\frac{\partial}{\partial W} \log p(Y, Z; W) \right] p(Y, Z; W) dZ \nonumber  \\
%&&= \int \left[\frac{\partial}{\partial W} \log p(Y, Z; W) \right] \frac{p(Y, Z; W)}{p(Y; W)} dZ  \nonumber \\
&&= \E_{p(Z|Y, W)} \left[\frac{\partial}{\partial W} \log p(Y, Z; W) \right]. \label{eq:cot}
\end{eqnarray}
The expectation with respect to $p(Z|Y, W)$ can be approximated by drawing samples from $p(Z|Y, W)$ and then computing the Monte Carlo average. 

The Langevin dynamics for sampling $Z \sim p(Z|Y, W)$ iterates 
\begin{eqnarray}
&&Z_{\tau + 1} = Z_\tau +   s U_\tau + \nonumber \\
&&\frac{s^2}{2}\left[\frac{1}{\sigma^2}( Y-f(Z_\tau; W))  \frac{\partial}{\partial Z} f(Z_\tau; W)  - {Z_\tau}\right],  
\label{eq:Langevin}
\end{eqnarray} 
where $\tau$ denotes the time step for the Langevin sampling, $s$ is the step size, and $U_\tau$ denotes a random vector that follows ${\rm N}(0, I_d)$.  The Langevin dynamics (\ref{eq:Langevin}) is an explain-away process, where the latent factors in $Z$ compete to explain away the current residual $Y - f(Z_\tau; W)$. 

To explain Langevin dynamics, its continuous time version  for sampling $\pi(x) \propto \exp[-{\cal E}(x)]$ is $x_{t + \Delta t} = x_t - \Delta t {\cal E}'(x_t)/2 + \sqrt{\Delta t} U_t$. The dynamics has $\pi$ as its stationary distribution, because it can be shown that for any well-behaved testing function $h$,  if $x_t \sim \pi$, then $\E[h(x_{t+\Delta t})] - \E[h(x_t)] \rightarrow 0$, as $\Delta t \rightarrow 0$, so that $x_{t+\Delta t} \sim \pi$. Alternatively, given $x_{t} = x$, suppose $x_{t+\Delta t} \sim K(x, y)$, then $[\pi(y) K(y, x)]/[\pi(x) K(x, y)] \rightarrow 1$ as $\Delta t \rightarrow 0$. 

The stochastic gradient algorithm of \citep{younes1999convergence} can be used for learning, where in each iteration, for each $Z_i$, only a single copy of $Z_i$ is sampled from $p(Z_i|Y_i, W)$ by running a finite number of steps of Langevin dynamics starting from the current value of $Z_i$, i.e., the warm start. With $\{Z_i\}$ sampled in this manner, we can update the parameter $W$ based on the gradient ${L}'(W)$, whose Monte Carlo approximation is: 
\begin{eqnarray} 
{L}'(W) &\approx& \sum_{i=1}^{n}  \frac{\partial}{\partial W} \log p(Y_i, Z_i; W) \nonumber\\
&=& - \sum_{i=1}^{n}   \frac{\partial}{\partial W}    \frac{1}{2\sigma^2} \|Y_i - f(Z_i; W)\|^2 \nonumber \\
&=&  \sum_{i=1}^{n}   \frac{1}{\sigma^2} (Y_i-f(Z_i; W)) \frac{\partial}{\partial W} f(Z_i; W). \label{eq:learning}
\end{eqnarray} 

Algorithm \ref{code:3} describes the details of the learning and sampling algorithm.  

\begin{algorithm}
	\caption{Alternating back-propagation}
	\label{code:3}
	\begin{algorithmic}[1]
		
		\REQUIRE ~~\\
		(1)  training examples $\{Y_i, i=1,...,n\}$ \\
		(2) number of Langevin steps $l$\\
		(3) number of learning iterations $T$
		
		\ENSURE~~\\
		(1) learned parameters $W$\\
		(2) inferred latent factors $\{Z_i, i =  1, ..., n\}$ 
		
		\item[]
		\STATE Let $t\leftarrow 0$, initialize $W$.
		\STATE Initialize $Z_i$, for $i =  1, ..., {n}$. 
		\REPEAT 
		\STATE {\bf Inferential back-propagation}: For each $i$, run $l$ steps of Langevin dynamics to sample $Z_i \sim p(Z_i|Y_i, W)$ with warm start, i.e., starting from the current $Z_i$, each step 
		follows equation (\ref{eq:Langevin}). 
		\STATE {\bf Learning back-propagation}: Update $W \leftarrow W + \gamma_t {L}'(W) $,  where ${L}'(W)$ is computed according to equation (\ref{eq:learning}), with learning rate $\gamma_t$. 
		\STATE Let $t \leftarrow t+1$
		\UNTIL $t = T$
	\end{algorithmic}
\end{algorithm}

If the Gaussian noise $U_\tau$ in the Langevin dynamics (\ref{eq:Langevin})  is removed, then the above algorithm becomes the alternating gradient descent algorithm.  It is possible to update both $W$ and $\{Z_i\}$ simultaneously by joint gradient descent.

Both the inferential back-propagation and the learning back-propagation  are guided by the residual  $Y_i - f(Z_i; W)$.
The inferential back-propagation is based on $\partial f(Z; W)/\partial Z$, whereas the learning back-propagation is based on $\partial f(Z; W)/\partial W$. Both gradients can be efficiently computed by back-propagation. The computations of the two gradients share most of their steps. Specifically, for the top-down ConvNet defined by (\ref{eq:Conv2}),  $\partial f(Z; W)/\partial W$ and $\partial f(Z; W)/\partial Z$ share the same code for the chain rule computation of $\partial Z^{(l-1)}/\partial Z^{(l)}$ for $l = 1, ..., L$.  Thus, the code for $\partial f(Z; W)/\partial Z$ is part of the code for $\partial f(Z; W)/\partial W$. 

In Algorithm \ref{code:3}, the Langevin dynamics samples from a gradually changing posterior distribution $p(Z_i|Y_i, W)$ because $W$ keeps changing. The updating of both $Z_i$ and $W$ collaborate to reduce the reconstruction error $\|Y_i - f(Z_i; W)||^2$.  The parameter $\sigma^2$  plays the role of annealing or tempering in Langevin sampling. If $\sigma^2$ is very large, then the posterior is close to the prior ${\rm N}(0, I_d)$. If $\sigma^2$ is very small, then the posterior may be multi-modal, but the evolving energy landscape  of  $p(Z_i|Y_i, W)$ may help alleviate the trapping of the local modes. In practice, we tune the value of $\sigma^2$ instead of estimating it. 
The Langevin dynamics can be extended to Hamiltonian Monte Carlo \citep{neal2011mcmc} or more sophisticated versions \citep{girolami2011riemann}.

%Alternatively, if Gaussian noise is added to the learning step yielding Langevin dynamics  in a full Bayesian treatment \citep{welling2011bayesian}, then the algorithm will become an alternating Langevin dynamics that samples the joint posterior distribution of both $\{Z_i\}$ and $W$. 

\section{Experiments}

The code in our experiments is based on the MatConvNet package of \citep{matconvnn}.  
%The data, code and more results including sounds and videos are available from the project page \url{http://www.stat.ucla.edu/~ywu/ABP/main.html}. 

The training images and sounds are scaled so that the intensities are within the range $[-1, 1]$.  We adopt the structure of the generator network of  \citep{radford2015unsupervised,Alexey2015}, where the top-down network consists of multiple layers of deconvolution by linear superposition, ReLU non-linearity, and up-sampling, with tanh non-linearity at the bottom-layer \citep{radford2015unsupervised} to make the signals fall within $[-1, 1]$. We also adopt batch normalization \citep{Ioffe2015BatchNA}. 

We fix $\sigma = .3$ for the standard deviation of the noise vector $\epsilon$. We use $l = 10$ or 30 steps of Langevin dynamics within each learning iteration, and the Langevin step size $s$  is set at .1 or $.3$. We run $T= 600$ learning iterations, with learning rate .0001, and momentum .5.  The learning algorithm produces the learned network parameters $W$ and the inferred latent factors $Z$ for each signal $Y$ in the end. The synthesized signals are obtained by $f(Z; W)$, where $Z$ is sampled from the prior distribution ${\rm N}(0, I_d)$. 

\subsection{Qualitative experiments}

\begin{figure}[h]
	\begin{center}
		\includegraphics[width=.15\linewidth]{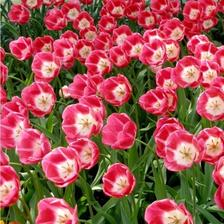}
		\includegraphics[width=.3\linewidth]{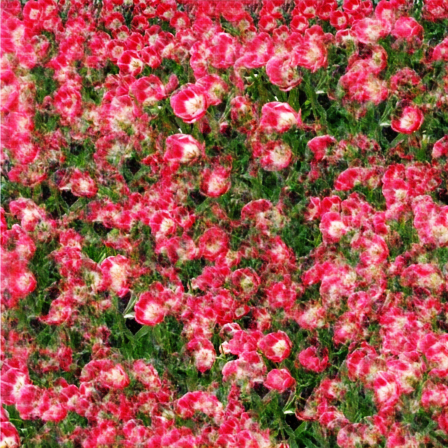}
		\includegraphics[width=.15\linewidth]{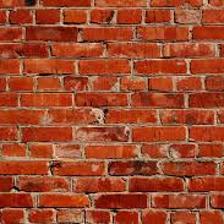}
		\includegraphics[width=.3\linewidth]{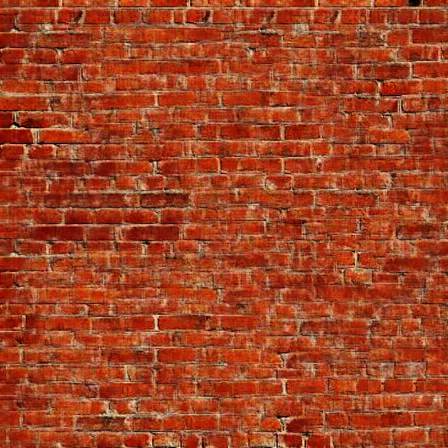}\\ \vspace{1mm}
		\includegraphics[width=.15\linewidth]{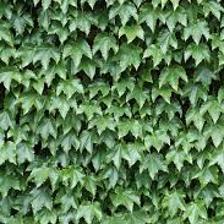}
		\includegraphics[width=.3\linewidth]{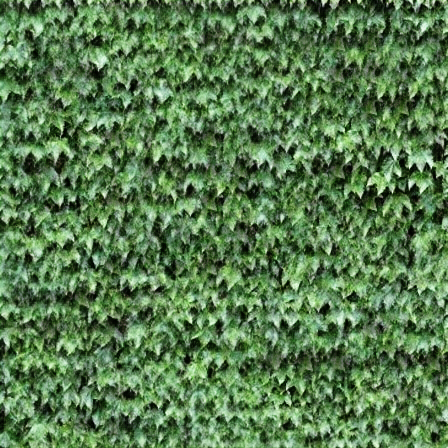}
		\includegraphics[width=.15\linewidth]{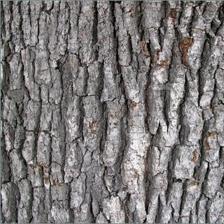}
		\includegraphics[width=.3\linewidth]{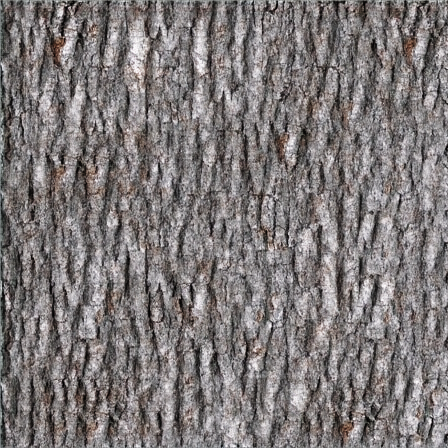}\\
		%\vspace{1mm}	
		\caption{Modeling texture patterns. For each example, {\em Left:} the $224 \times 224$  observed image. {\em Right:} the $448 \times 448$ generated image. }
		\label{fig:texture1}
	\end{center}
\end{figure}

\indent\indent{\em Experiment 1. Modeling texture patterns}.  We learn a separate model from each texture image. The images are collected from the Internet, and then resized to 224$\times$ 224. The synthesized images are 448 $\times$ 448. Figures \ref{fig:texture1} shows four examples.  

The factors $Z$ at the top layer form a  $\sqrt{d} \times \sqrt{d}$ image, with each pixel following ${\rm N}(0, 1)$ independently. The  $\sqrt{d} \times \sqrt{d}$  image $Z$ is then transformed to $Y$ by the top-down ConvNet.  We use $d= 7^2$ in the learning stage for all the texture experiments. In order to obtain the synthesized image,  we randomly sample a 14 $\times$ 14 $Z$ from N$(0, I)$, and then expand the learned network $W$ to generate the 448 $\times$ 448 synthesized image $f(Z; W)$. 

The training network is as follows. Starting from $7 \times 7$  image $Z$, the network has 5 layers of deconvolution with $5 \times 5$ kernels (i.e., linear superposition of $5 \times 5$ basis functions), with an up-sampling factor of 2 at each layer (i.e., the basis functions are 2 pixels apart). The number of channels in the first layer is 512 (i.e., 512 translation invariant basis functions), and is decreased by a factor 2 at each layer. The Langevin steps $l = 10$ with step size $s = .1$.

\begin{figure}[h]
	\begin{center}
		\includegraphics[width=.95\linewidth]{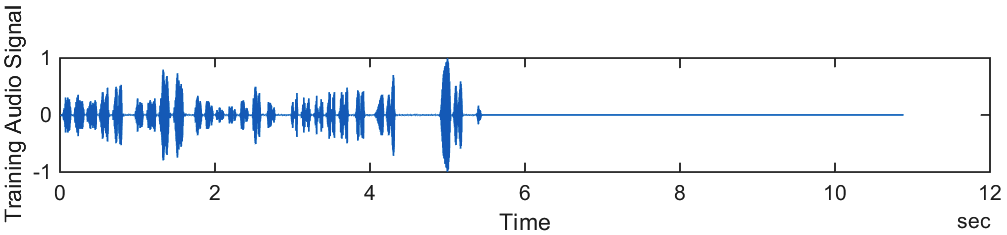}
		\includegraphics[width=.95\linewidth]{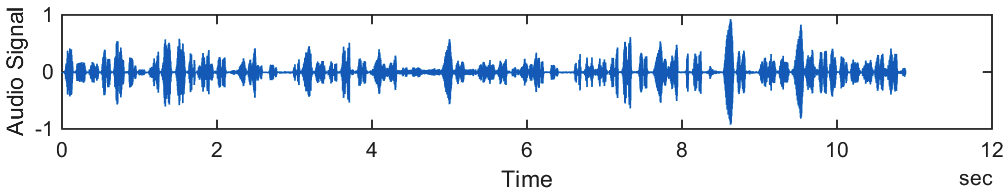}
		\caption{Modeling sound patterns. {\em Row 1:} the waveform of the training sound (the range is 0-5 seconds). {\em Row 2:} the waveform of the synthesized sound (the range is 0-11 seconds). }
		\label{fig:sound}
	\end{center}
\end{figure}

{\em Experiment  2. Modeling sound patterns}.  A sound signal can be treated as a one-dimensional texture image \citep{mcdermott2011sound}. The sound data are collected from the Internet. Each training signal is a 5 second clip with the sampling rate of 11025 Hertz and is represented as a $1 \times 60000$ vector.  We learn a separate model from each sound signal. 

The latent factors $Z$ form a sequence that follows N$(0, I_d)$,  with $d=6$.  The top-down network consists of  4  layers of deconvolution with kernels of size $1 \times 25$, and up-sampling factor of 10. The number of channels in the first layer is 256, and decreases by a factor of 2 at each layer.  For synthesis, we start from a longer Gaussian white noise sequence $Z$ with $d = 12$ and generate the synthesized sound by expanding the learned network. Figure \ref{fig:sound} shows the waveforms of the observed sound signal in the first row and the synthesized sound signal in the second row. 
%The reader can listen to the sounds at  \url{http://www.stat.ucla.edu/~ywu/ABP/main.html}. 

\begin{figure}[h]
	\begin{center}
		\includegraphics[width=.48\linewidth]{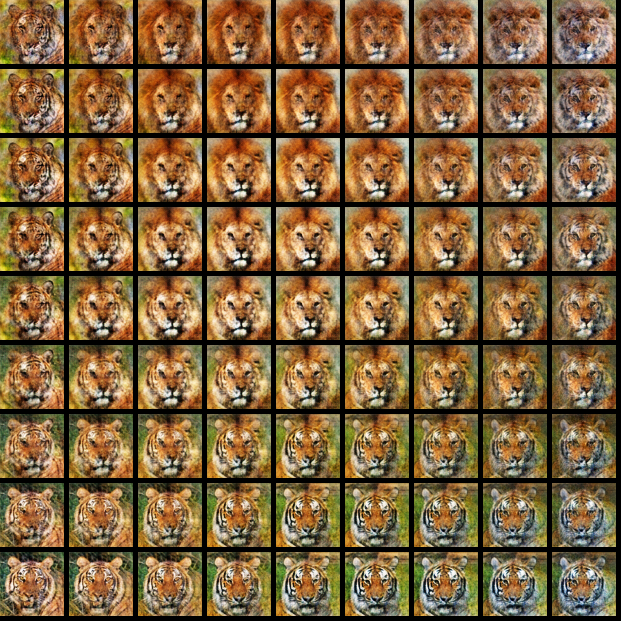} 
		\includegraphics[width=.48\linewidth]{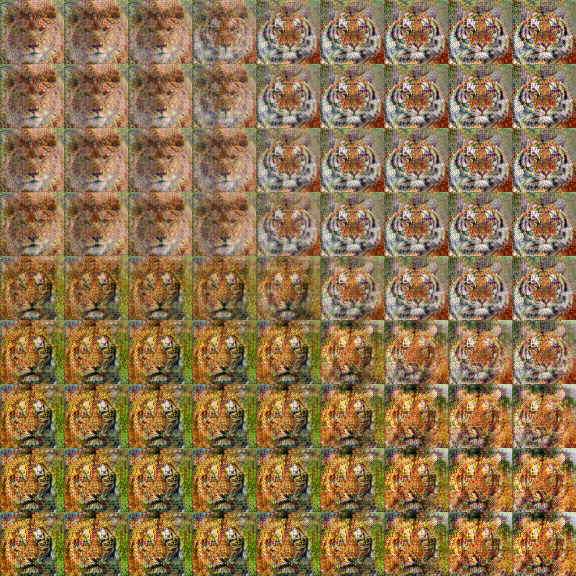}
		\caption{Modeling object patterns. {\em Left:} the synthesized images generated by our method. They are generated by $f(Z; W)$ with the learned $W$, where $Z = (z_1, z_2)  \in [-2, 2]^2$, and $Z$ is discretized into $9 \times 9$ values. {\em Right:} the synthesized images generated using Deep Convolutional Generative Adversarial Net (DCGAN). $Z$ is discretized into $9 \times 9$ values within $[-1, 1]^2$.   }
		\label{fig:lion-tiger}
	\end{center}
\end{figure}

\begin{figure}[h]
	\begin{center}
		\includegraphics[width=.32\linewidth]{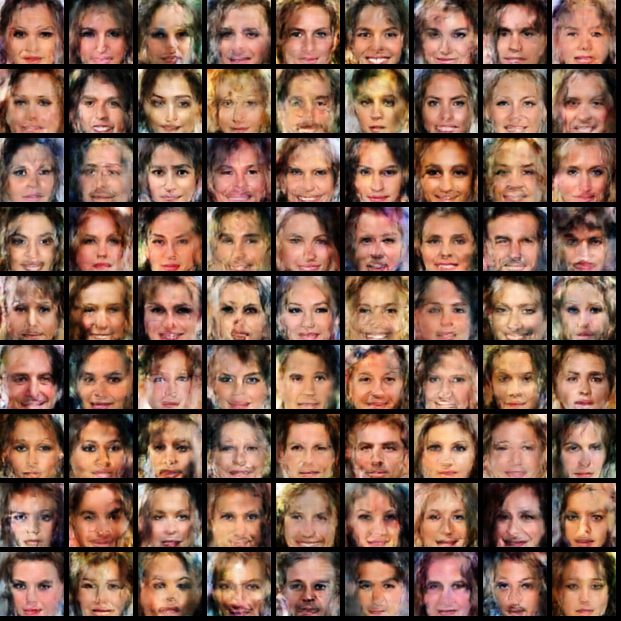} 
		\includegraphics[width=.32\linewidth]{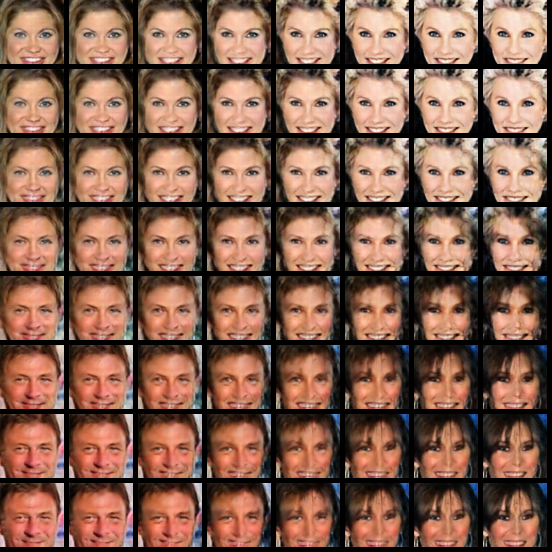} 
		\includegraphics[width=.32\linewidth]{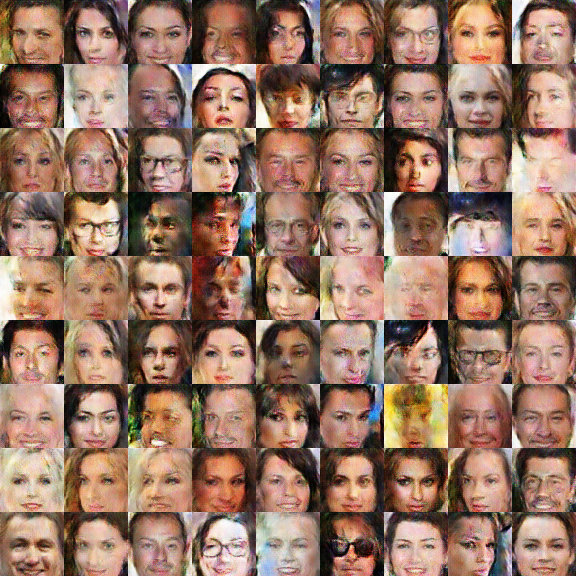}
		\caption{Modeling object patterns. {\em Left:}  each  image generated by our method is obtained by first sampling $Z \sim {\rm N}(0, I_{100})$ and then generating the image by $f(Z; W)$ with the learned $W$. {\em Middle:} interpolation. The images at the four corners are reconstructed from the inferred $Z$ vectors of four images randomly selected from the training set. Each image in the middle is obtained by first interpolating the $Z$ vectors of the four corner images, and then generating the image by $f(Z; W)$. {\em Right:} the synthesized images generated by DCGAN, where $Z$ is a $100$ dimension vector sampled from uniform distribution. }
		\label{fig:face1000}
	\end{center}
\end{figure}

{\em Experiment  3. Modeling object patterns}. We model object patterns using the network structure that is essentially the same as the network for the texture model, except that we include a fully connected layer under the latent factors $Z$, now a $d$-dimensional vector. The images are $64 \times 64$. We use ReLU with a leaking factor .2 \citep{maas2013rectifier,Xu2015EmpiricalEO}.  The Langevin steps $l = 30$ with step size $s = .3$.

In the first experiment, we learn a model where $Z$ has two components, i.e., $Z = (z_1, z_2)$, and $d = 2$. The training data are 11 images of 6 tigers and 5 lions. After training the model, we generate images using the learned top-down ConvNet for $(z_1, z_2) \in [-2, 2]^2$, where we discretize both $z_1$ and $z_2$ into 9 equally spaced values. The left panel of Figure \ref{fig:lion-tiger}  displays the synthesized images on the $9 \times 9$ panel. 

In the second experiment,  we learn a model with $d = 100$ from 1000 face images randomly selected from the CelebA dataset \citep{liu2015deep}. The left panel of Figure \ref{fig:face1000} displays the images generated by the learned model. The middle panel displays the interpolation results. The images at the four corners are generated by the $Z$ vectors of four images  randomly selected from the training set. The images in the middle are obtained by first interpolating the $Z$'s of the four corner images using the sphere interpolation \citep{Dinh2016DensityEU} and then generating the images by the learned ConvNet. 

We also provide qualitative comparison with Deep Convolutional Generative Adversarial Net (DCGAN) \citep{goodfellow2014generative,radford2015unsupervised}. 
The right panel of Figure \ref{fig:lion-tiger} shows the generated results for the lion-tiger dataset using $2$-dimensional $Z$. The right panel of Figure \ref{fig:face1000} displays the generated results trained on 1000 aligned faces from celebA dataset, with $d = 100$. We use the code from {\small \url{https://github.com/carpedm20/DCGAN-tensorflow}},  with the tuning parameters as in \citep{radford2015unsupervised}. We run $T = 600$ iterations as in our method. 

\begin{figure}[h]
	\begin{center}
		\includegraphics[width=.11\linewidth]{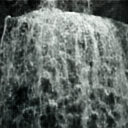}
		\includegraphics[width=.11\linewidth]{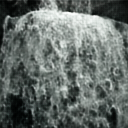}
		\includegraphics[width=.11\linewidth]{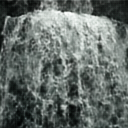}
		\includegraphics[width=.11\linewidth]{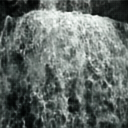}
		\includegraphics[width=.11\linewidth]{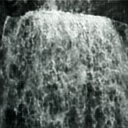}
		\includegraphics[width=.11\linewidth]{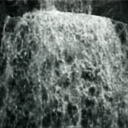}
		\includegraphics[width=.11\linewidth]{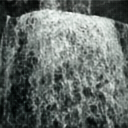}\\	\vspace{.5mm}
		\includegraphics[width=.11\linewidth]{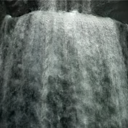}
		\includegraphics[width=.11\linewidth]{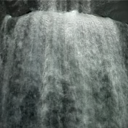}
		\includegraphics[width=.11\linewidth]{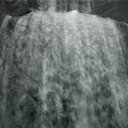}
		\includegraphics[width=.11\linewidth]{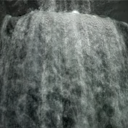}
		\includegraphics[width=.11\linewidth]{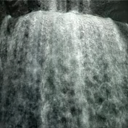}
		\includegraphics[width=.11\linewidth]{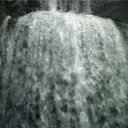}
		\includegraphics[width=.11\linewidth]{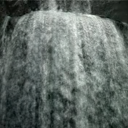}\\	\vspace{.5mm}
		\includegraphics[width=.11\linewidth]{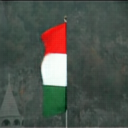}
		\includegraphics[width=.11\linewidth]{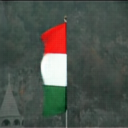}
		\includegraphics[width=.11\linewidth]{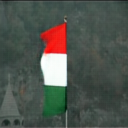}
		\includegraphics[width=.11\linewidth]{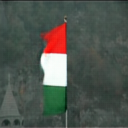}
		\includegraphics[width=.11\linewidth]{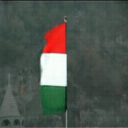}
		\includegraphics[width=.11\linewidth]{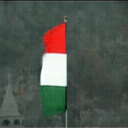}
		\includegraphics[width=.11\linewidth]{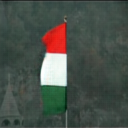}\\	\vspace{.5mm}
		\includegraphics[width=.11\linewidth]{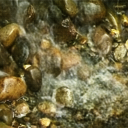}
		\includegraphics[width=.11\linewidth]{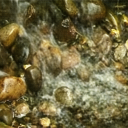}
		\includegraphics[width=.11\linewidth]{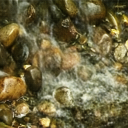}
		\includegraphics[width=.11\linewidth]{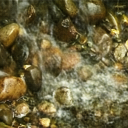}
		\includegraphics[width=.11\linewidth]{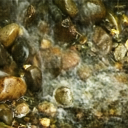}
		\includegraphics[width=.11\linewidth]{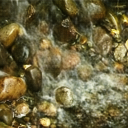}
		\includegraphics[width=.11\linewidth]{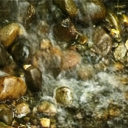}
		\caption{Modeling dynamic textures. {\em Row 1:} a segment of the synthesized sequence by our method. {\em Row 2:} a sequence by the method of \citep{dorettoCWS03}. {\em Rows 3 and 4:}  two more sequences by our method. }
		\label{fig:dynamicSequence}
	\end{center}
\end{figure}

{\em Experiment  4. Modeling dynamic patterns}.  We model a textured motion \citep{Wang03modelingtextured} or a dynamic texture \citep{dorettoCWS03} by a non-linear dynamic system $Y_t = f(Z_t; W) + \epsilon_t$, and $Z_{t+1} = A Z_t + \eta_t$, where we assume the latent factors follow a vector auto-regressive model, where $A$ is a $d \times d$ matrix, and $\eta_t \sim {\rm N}(0, Q)$ is the innovation. This model is a direct generalization of the linear dynamic system of \citep{dorettoCWS03},  where $Y_t$ is reduced to $Z_t$ by principal component analysis (PCA) via singular value decomposition (SVD). We learn the model in two steps. (1) Treat $\{Y_t\}$ as independent examples and learn $W$ and infer $\{Z_t\}$ as before. (2) Treat $\{Z_t\}$ as the training data, learn $A$ and $Q$ as in \citep{dorettoCWS03}. After that, we can synthesize a new dynamic texture. We start from $Z_0 \sim {\rm N}(0, I_d)$, and then generate the sequence according to the learned model (we discard a burn-in period of 15 frames).  Figure \ref{fig:dynamicSequence} shows some experiments, where we set $d=20$. The first row is a segment of the sequence generated by our model, and the second row is generated by the method of  \citep{dorettoCWS03}, with the same dimensionality of $Z$. It is possible to generalize the auto-regressive model of $Z_t$ to recurrent network. We may also treat the video sequences as 3D images, and learn generator networks with 3D spatial-temporal filters or basis functions. 

\subsection{Quantitative experiments}

\indent\indent{\em Experiment  5. Learning from incomplete data}. Our method can learn from images with occluded pixels. This task is inspired by the fact that most of the images contain occluded objects.  It can be considered a non-linear generalization of matrix completion in recommender system.

Our method can be adapted to this task with minimal modification. The only modification involves the computation of $\|Y - f(Z; W)\|^2$. For a  fully observed image, it is computed by summing over all the pixels. For a partially observed image, we compute it by summing over only the observed pixels. Then we can continue to use the alternating back-propagation algorithm to infer $Z$ and learn $W$.  With inferred $Z$ and learned $W$, the image can be automatically recovered by $f(Z; W)$. In the end, we will be able to accomplish the following tasks: (T1) Recover the occluded pixels of  training images. (T2) Synthesize new images from the learned model.  (T3) Recover the occluded pixels of testing images using the learned model.

%\bgroup
%\def\arraystretch{1.2}
\begin{table}[h]
	\begin{center}
		\begin{tabular}{|c|c|c|c|c|c|}
			\hline experiment			&	P.5	& P.7		& P.9		&	M20 	& M30	\\
			\hline error	&	.0571	& .0662	& .0771	& 	.0773 	& .1035\\
			\hline 
		\end{tabular}		
		\caption{Recovery errors in 5 experiments of learning from occluded images.} 
		\label{tab:face}
	\end{center}
\end{table}
%\egroup

\begin{figure}[h]
	\centering
	\subfloat{
		\includegraphics[width=.088\linewidth]{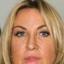}
		\includegraphics[width=.088\linewidth]{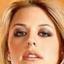}
		\includegraphics[width=.088\linewidth]{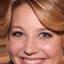}
		\includegraphics[width=.088\linewidth]{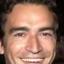}
		\includegraphics[width=.088\linewidth]{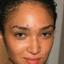}
		\includegraphics[width=.088\linewidth]{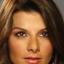}
		\includegraphics[width=.088\linewidth]{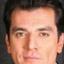}
		\includegraphics[width=.088\linewidth]{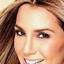}
		\includegraphics[width=.088\linewidth]{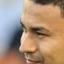}
		\includegraphics[width=.088\linewidth]{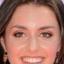}
	}\\[1px]
	
	\subfloat{
		\includegraphics[width=.088\linewidth]{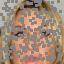}
		\includegraphics[width=.088\linewidth]{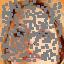}
		\includegraphics[width=.088\linewidth]{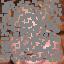}
		\includegraphics[width=.088\linewidth]{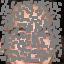}
		\includegraphics[width=.088\linewidth]{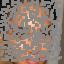}
		\includegraphics[width=.088\linewidth]{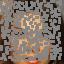}
		\includegraphics[width=.088\linewidth]{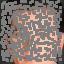}
		\includegraphics[width=.088\linewidth]{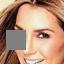}
		\includegraphics[width=.088\linewidth]{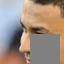}	
		\includegraphics[width=.088\linewidth]{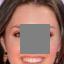}	
	}\\[1px]
	
	\subfloat{
		\includegraphics[width=.088\linewidth]{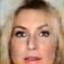}
		\includegraphics[width=.088\linewidth]{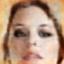}
		\includegraphics[width=.088\linewidth]{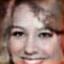}
		\includegraphics[width=.088\linewidth]{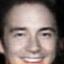}
		\includegraphics[width=.088\linewidth]{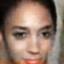}
		\includegraphics[width=.088\linewidth]{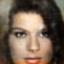}
		\includegraphics[width=.088\linewidth]{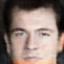}
		\includegraphics[width=.088\linewidth]{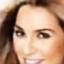}
		\includegraphics[width=.088\linewidth]{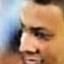}
		\includegraphics[width=.088\linewidth]{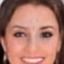}	
	}	
	\caption{Learning from incomplete data. The 10 columns belong to experiments P.5, P.7, P.9, P.9, P.9, P.9, P.9, M20, M30, M30 respectively.  {\em Row 1:}  original images, not observed in learning. {\em Row 2:}  training images. {\em Row 3:}  recovered images during learning. }
	\label{fig:face}
\end{figure}

We want to emphasize that in our experiments, all the training images are partially occluded. Our experiments are different from (1) de-noising auto-encoder \citep{vincent2008extracting}, where the training images are fully observed, and noises are added as a matter of regularization, (2) in-painting or de-noising, where the prior model or regularization has already been learned or given. (2) is about task (T3) mentioned above, but not about tasks (T1) and (T2). 

Learning from incomplete data can be difficult for GAN and VAE, because the occluded pixels are different for different training images.

We evaluate our method on 10,000  images randomly selected from CelebA dataset. We design 5 experiments, with two types of occlusions: (1) 3 experiments are about salt and pepper occlusion, where we randomly place $3 \times 3$ masks on the $64 \times 64$ image domain to cover roughly 50\%, 70\% and 90\% of pixels respectively. These 3 experiments are denoted P.5, P.7, and P.9 respectively (P for pepper). (2) 2 experiments are about single region mask occlusion, where we randomly place a $20 \times 20$ or $30 \times 30$ mask on the $64 \times 64$ image domain. These 2 experiments are denoted M20 and M30 respectively (M for mask). We set $d = 100$. 
Table~\ref{tab:face} displays the recovery errors of the 5 experiments, where the error is defined as per pixel difference (relative to the range of the pixel values) between the original image and the recovered image on the occluded  pixels.  We emphasize that the recovery errors are not training errors, because the intensities of the occluded pixels are not observed in training. 
 Figure~\ref{fig:face} displays recovery results.
 In experiment P.9, 90$\%$ of pixels are occluded, but we can still learn the model and recover the original images.

%\bgroup
%\def\arraystretch{1.2}
\begin{table}[h]
	\begin{center}
		\begin{tabular}{|c|c|c|c|}
			\hline 	experiment		&	$d=20$	& $d=60$	& $d=100$		\\
			\hline 	error	&	.0795	& .0617	& .0625	\\
			\hline 
		\end{tabular}		
		\caption{Recovery errors in 3 experiments of learning from compressively sensed images.} 
		\label{tab:cs}
	\end{center}
\end{table}
%\egroup

\begin{figure}[h]
	\centering
	\subfloat{
		\includegraphics[width=.088\linewidth]{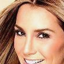}
		\includegraphics[width=.088\linewidth]{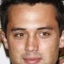}
		\includegraphics[width=.088\linewidth]{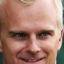}
		\includegraphics[width=.088\linewidth]{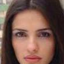}
		\includegraphics[width=.088\linewidth]{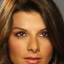}
		\includegraphics[width=.088\linewidth]{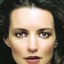}
		\includegraphics[width=.088\linewidth]{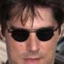}
		\includegraphics[width=.088\linewidth]{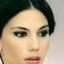}
		\includegraphics[width=.088\linewidth]{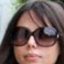}
		\includegraphics[width=.088\linewidth]{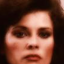}
	}\\[1px]
	
	\subfloat{
		\includegraphics[width=.088\linewidth]{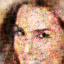}
		\includegraphics[width=.088\linewidth]{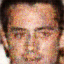}
		\includegraphics[width=.088\linewidth]{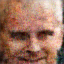}
		\includegraphics[width=.088\linewidth]{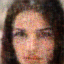}
		\includegraphics[width=.088\linewidth]{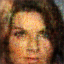}
		\includegraphics[width=.088\linewidth]{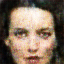}
		\includegraphics[width=.088\linewidth]{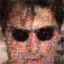}
		\includegraphics[width=.088\linewidth]{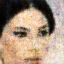}
		\includegraphics[width=.088\linewidth]{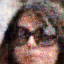}
		\includegraphics[width=.088\linewidth]{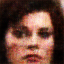}
	}	
	\caption{Learning from indirect data. {\em Row 1:} the original $64 \times 64 \times 3$  images, which are projected onto 1,000 white noise images. {\em Row 2:} the recovered images during learning.}
	\label{fig:cs}
\end{figure}

{\em Experiment  6. Learning from indirect data}. We can  learn the model from the compressively sensed data  \citep{candes2006robust}. We generate a set of white noise images as random projections. We then project the training images on these white noise images. We can  learn the model from the random projections  instead of the original images. We only need to replace $\|Y - f(Z; W)\|^2$ by $\|SY - S f(Z; W)\|^2$, where $S$ is the given white noise sensing matrix, and $SY$ is the observation. We can treat $S$ as a fully connected layer of known filters below $f(Z; W)$, so that we can continue to use alternating back-propagation to infer $Z$ and learn $W$, thus recovering the image by $f(Z; W)$.  In the end, we will be able to (T1) Recover the original images from their  projections during learning. (T2) Synthesize new images from the learned model. (T3) Recover testing images from their projections based on the learned model. Our experiments are different from traditional compressed sensing, which is task (T3), but not tasks (T1) and (T2).  Moreover, the image recovery in our work is based on non-linear dimension reduction instead of linear sparsity. 

We evaluate our method on 1000 face images randomly selected from CelebA dataset.  These images are projected onto $K = 1000$ white noise images with each pixel randomly sampled from ${\rm N}(0, .5^2)$. After this random projection, each image of size $64\times64\times3$ becomes a $K$-dimensional vector.  We show the recovery errors for different latent dimensions $d$ in Table~\ref{tab:cs}, where the recovery error is defined as the per pixel difference (relative to the range of the pixel values) between the original image and the recovered image.  
%We again emphasize that the recovery errors are not training errors, because the original face images are not observed during our learning process. 
Figure~\ref{fig:cs} shows some recovery results. 

\begin{figure}
	\centering
	\subfloat{
		\includegraphics[width=.088\linewidth]{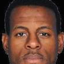}
		\includegraphics[width=.088\linewidth]{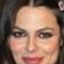}
		\includegraphics[width=.088\linewidth]{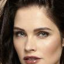}
		\includegraphics[width=.088\linewidth]{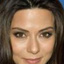}
		\includegraphics[width=.088\linewidth]{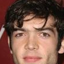}
		\includegraphics[width=.088\linewidth]{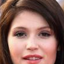}
		\includegraphics[width=.088\linewidth]{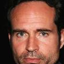}
		\includegraphics[width=.088\linewidth]{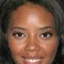}
		\includegraphics[width=.088\linewidth]{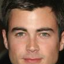}
		\includegraphics[width=.088\linewidth]{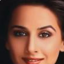}
	}\\[1px]
	
	\subfloat{
		\includegraphics[width=.088\linewidth]{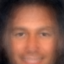}
		\includegraphics[width=.088\linewidth]{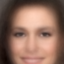}
		\includegraphics[width=.088\linewidth]{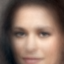}
		\includegraphics[width=.088\linewidth]{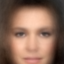}
		\includegraphics[width=.088\linewidth]{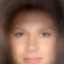}
		\includegraphics[width=.088\linewidth]{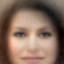}
		\includegraphics[width=.088\linewidth]{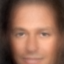}
		\includegraphics[width=.088\linewidth]{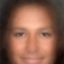}
		\includegraphics[width=.088\linewidth]{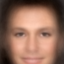}
		\includegraphics[width=.088\linewidth]{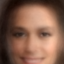}
	}\\[1px]
	
	\subfloat{
		\includegraphics[width=.088\linewidth]{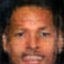}
		\includegraphics[width=.088\linewidth]{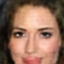}
		\includegraphics[width=.088\linewidth]{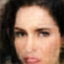}
		\includegraphics[width=.088\linewidth]{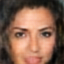}
		\includegraphics[width=.088\linewidth]{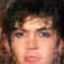}
		\includegraphics[width=.088\linewidth]{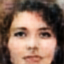}
		\includegraphics[width=.088\linewidth]{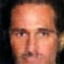}
		\includegraphics[width=.088\linewidth]{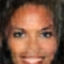}
		\includegraphics[width=.088\linewidth]{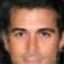}
		\includegraphics[width=.088\linewidth]{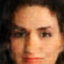}
	}	
	\caption{Comparison between our method and PCA.  {\em Row 1:}  original testing images. {\em Row 2:} reconstructions by PCA eigenvectors learned from training images. {\em Row 3: }  reconstructions  by the generator learned from training images. $d = 20$ for both methods. }
	\label{fig:comp}
\end{figure}

{\em Experiment  7. Model evaluation by reconstruction error on testing data}. After learning the model from the training images (now assumed to be fully observed), we can evaluate the model by the reconstruction error on the testing images. We randomly select 1000 face images for training and 300 images for testing from CelebA dataset.  After learning, we infer the latent factors $Z$ for each testing image using inferential back-propagation, and then reconstruct the testing image by $f(Z; W)$ using the inferred $Z$ and the learned $W$. In the inferential back-propagation for inferring $Z$,  we initialize  $Z \sim {\rm N}(0, I_d)$, and run 300 Langevin steps with step size .05. Table~\ref{tab:PCA} shows the reconstruction errors of alternating back-propagation learning (ABP) as compared to PCA learning for different latent dimensions $d$. Figure~\ref{fig:comp} shows some reconstructed testing images. For PCA, we learn the $d$ eigenvectors from the training images, and then project the testing images on the learned eigenvectors for reconstruction.

%\bgroup
%\def\arraystretch{1.2}
\begin{table}
	\begin{center}
		\begin{tabular}{|c|c|c|c|c|}
			\hline    experiment 	& $d=20$	& $d=60$ & $d=100$ & $d=200$	\\
			\hline 	ABP		&	.0810 	& .0617 & .0549 & .0523	\\
			\hline  PCA 	&   .1038 	& .0820 & .0722 & .0621		\\ 		
			\hline 
		\end{tabular}		
		\caption{Reconstruction errors on testing images, after learning from training images using our method (ABP) and PCA. } 
		\label{tab:PCA}
	\end{center}
\end{table}
%\egroup

Experiments 5-7 may be used to evaluate generative models in general. Experiments 5 and 6 appear new, and we have not found comparable methods that can accomplish all three tasks (T1), (T2), and (T3) simultaneously. 

\section{Conclusion}

This paper proposes an alternating back-propagation algorithm for training the generator network. We recognize that the generator network is a non-linear generalization of the factor analysis model, and develop the alternating back-propagation algorithm as the non-linear generalization of the alternating regression scheme of the Rubin-Thayer EM algorithm for fitting the factor analysis model. The alternating back-propagation algorithm iterates  the inferential back-propagation for inferring the latent factors and the learning back-propagation for updating the parameters. Both back-propagation steps share most of their computing steps in the chain rule calculations. 

Our learning algorithm is perhaps the most canonical algorithm for training the generator network. It is based on maximum likelihood, which is theoretically the most accurate estimator. The maximum likelihood learning seeks to explain and charge the whole dataset uniformly, so that there is little concern of under-fitting or biased fitting.  

As an unsupervised learning algorithm, the alternating back-propagation algorithm is a natural generalization of the original back-propagation algorithm for supervised learning. It adds an inferential back-propagation step to the learning back-propagation step, with minimal overhead in coding and affordable overhead in computing. The inferential back-propagation seeks to perform accurate explaining-away inference of the latent factors. It can be worthwhile for tasks such as learning from incomplete or indirect data, or learning models where the latent factors themselves follow sophisticated prior models with unknown parameters. The inferential back-propagation may also be used to evaluate the generators learned by other methods on tasks such as reconstructing or completing testing data. 

Our method or its variants can be applied to non-linear matrix factorization and completion. It can also be applied to problems where some components or aspects of the factors are supervised. 
%It can also be applied to models with multiple layers of latent factors. 

\section*{Code, images, sounds, and videos}

{\small \url{http://www.stat.ucla.edu/~ywu/ABP/main.html}}

\section*{Acknowledgement}

%We thank the three reviewers and the area chair for their insightful comments and helpful suggestions. 
We thank Yifei (Jerry) Xu for his help with the experiments during his 2016 summer visit. We thank Jianwen Xie for helpful discussions.   

The work is supported by NSF DMS 1310391, DARPA SIMPLEX N66001-15-C-4035, ONR MURI N00014-16-1-2007, and DARPA ARO W911NF-16-1-0579.

\section{Appendix} 

\subsection{ReLU  and piecewise  factor analysis} 

The generator network  is $Y = f(Z; W) + \epsilon$, $Z^{(l-1)} = f_l(W_l Z^{(l)} + b_l)$, $l = 1, ..., L$, with $Z^{(0)} = f(Z; W)$, and $Z^{(L)} = Z$. The element-wise non-linearity $f_l$ in modern ConvNet is usually the two-piece linearity, such as rectified linear unit (ReLU) \citep{krizhevsky2012imagenet} or the leaky ReLU \citep{maas2013rectifier,Xu2015EmpiricalEO}. Each ReLU unit corresponds to a binary switch. For the case of non-leaky ReLU, following the analysis of \citep{pascanu2013number}, we can write
$
Z^{(l-1)} = \delta_l (W_l Z^{(l)} + b_l), \label{eq:Conv3}
$ where $\delta_l = {\rm diag}(1(W_l Z^{(l)} + b_l > 0))$ is a diagonal matrix,  $1()$ is an element-wise indicator function. For the case of leaky ReLU,  the 0 values on the diagonal  are replaced by a leaking factor (e.g., .2). 

$\delta = (\delta_l, l = 1, ..., L)$ forms a classification of $Z$ according to the network $W$. Specifically, the factor space of $Z$ is divided into a large number of pieces by the hyperplanes $W_l Z^{(l)} + b_l = 0$, and each piece is indexed by an instantiation of $\delta$.  We can write $\delta = \delta(Z; W)$ to make explicit its dependence on $Z$ and $W$. On the piece indexed by $\delta$,  $f(Z; W) = W_{\delta} Z + b_\delta$. Assuming $b_l = 0, \forall l$, for simplicity, we have $W_\delta = \delta_1 W_1 ...\delta_L W_L$. Thus each piece defined by $\delta = \delta(Z; W)$ corresponds to a linear  factor analysis $Y = W_\delta Z + \epsilon$, whose basis $W_\delta$ is a multiplicative recomposition of the basis functions at multiple layers $(W_l, l = 1, ..., L)$, and the recomposition is controlled by the binary switches at multiple layers $\delta  = (\delta_l, l = 1, ..., L)$. Hence the top-down ConvNet amounts to a reconfigurable  basis $W_\delta$ for representing $Y$, and  the model is a piecewise linear factor analysis. If we retain the bias term, we will have $Y = W_\delta Z + b_\delta + \epsilon$, for an overall bias term that depends on $\delta$. So the distribution of $Y$ is essentially piecewise Gaussian. 

The generator model can be considered an explicit implementation of the local linear embedding \citep{roweis2000nonlinear}, where $Z$ is the embedding of $Y$. In local linear embedding, the mapping between $Z$ and $Y$ is implicit. In the generator model,  the mapping from $Z$ to $Y$ is explicit. With ReLU ConvNet, the mapping is piecewise linear, which  is consistent with local linear embedding,  except that the partition of the linear pieces by $\delta(Z; W)$  in the generator model is learned automatically.
%without resorting to a pre-defined neighborhood system in the high dimensional space of $Y$ as in local linear embedding.  %The generator model   is also related to the auto-encoder, where $Z$ is the encoding of $Y$, and $Y$ is the decoding of $Z$. 

The inferential back-propagation is a Langevin dynamics on the energy function $\|Y - f(Z; W)\|^2/(2\sigma^2) + \|Z\|^2/2$. With $f(Z; W) = W_\delta Z $, $\partial f(Z; W)/\partial Z = W_\delta$. If $Z$ belongs to the piece defined by $\delta$, then the inferential back-propagation seeks to approximate $Y$ by the basis  $W_\delta$ via a ridge regression. Because $Z$ keeps changing during the Langevin dynamics, $\delta(Z; W)$ may also be changing,  and the algorithm searches for the optimal reconfigurable basis $W_\delta$ to approximate $Y$. We may solve $Z$ by second-order methods such as iterated ridge regression, which can be computationally more expensive than the simple gradient descent. 

\subsection{EM, density mapping, and density shifting} 

Suppose the training data $\{Y_i, i = 1, ..., n\}$ come from a data distribution $\P(Y)$. To understand how the alternating back-propagation algorithm or its EM idealization maps the prior distribution of the latent factors $p(Z)$  to the data distribution $\P(Y)$ by the learned  $g(Z; W)$,  we define 
\begin{eqnarray}
\P(Z, Y; W) = \P(Y) p(Z|Y, W) \nonumber \\ 
= \P(Z; W) \P(Y|Z, W), \label{eq:dataprior}
\end{eqnarray}
 where $\P(Z; W) = \int p(Z|Y, W) \P(Y) dY$ is obtained by averaging the posteriors $p(Z|Y; W)$ over the observed data $Y \sim \P$. That is, $\P(Z; W)$ can be considered the data prior. 
The data prior $\P(Z; W)$ is close to the true prior $p(Z)$  in the sense that  
\begin{eqnarray}
\KL(\P(Z; W)|p(Z)) \leq \KL(\P(Y)|p(Y; W)) \label{eq:dataprior2}\\
= \KL(\P(Z, Y; W)|p(Z, Y; W)). \nonumber
\end{eqnarray}
%The difference between the two sides of (\ref{eq:dataprior2}) is $\KL(\P(Y|Z, W)|p(Y|Z, W))$. 
 The right hand side of (\ref{eq:dataprior2})  is minimized at the maximum likelihood estimate
 $\hat{W}$, hence the data prior $\P(Z; \hat{W})$ at $\hat{W}$ should be especially close to the true prior $p(Z)$. In other words, at $\hat{W}$, the posteriors $p(Z|Y, \hat{W})$ of all the data points $Y \sim \P$ tend to pave the true prior $p(Z)$.  
 
From Rubin's multiple imputation point of view \citep{rubin2004multiple} of the EM algorithm,  the E-step of EM infers $Z_i^{(m)} \sim p(Z_i|Y_i, W_t)$ for $m = 1, ...,M$, where $M$ is the number of multiple imputations or multiple guesses of $Z_i$. The multiple guesses  account for the uncertainty in inferring $Z_i$ from $Y_i$. The M-step of EM maximizes  $Q(W) = \sum_{i=1}^{n} \sum_{m=1}^{M} \log p(Y_i, Z_i^{(m)}; W)$ to obtain $W_{t+1}$.  For each data point $Y_i$, $W_{t+1}$ seeks to reconstruct $Y_i$ by $g(Z; {W})$ from the inferred latent factors $\{Z_i^{(m)}, m = 1, ..., M\}$. In other words, the M-step seeks to map $\{Z_i^{(m)}\}$ to $Y_i$. Pooling over all $i = 1, ..., n$,  $\{Z_i^{(m)}, \forall i, m\} \sim \P(Z; W_t)$,  hence the M-step seeks to map $\P(Z; W_t)$ to the data distribution $\P(Y)$. Of course the mapping from $\{Z_i^{(m)}\}$ to  $Y_i$ cannot be exact. In fact, $g(Z; W)$ maps $\{Z_i^{(m)}\}$  to a $d$-dimensional patch around the $D$-dimensional $Y_i$. The local patches for all $\{Y_i, \forall i\}$ patch up the $d$-dimensional manifold form by the $D$-dimensional observed examples and their interpolations. The EM algorithm is a process of density shifting, so that $\P(Z; W)$ shifts towards $p(Z)$, thus $g(Z; W)$ maps $p(Z)$ to $\P(Y)$.

\subsection{Factor analysis and alternating  regression} 

The alternating back-propagation algorithm is inspired by Rubin-Thayer EM algorithm for factor analysis,  where both the observed data model $p(Y|W)$ and the posterior distribution $p(Z|Y, W)$ are available in closed form. The EM algorithm for factor analysis can be interpreted as alternating linear regression  \citep{rubin1982algorithms,liu1998parameter}.  

In the factor analysis model $Z \sim {\rm N}(0, I_d)$, $Y = W Z + \epsilon$, $\epsilon \sim {\rm N}(0, \sigma^2 I_D)$. The joint distribution of $(Z, Y)$ is 
\begin{eqnarray}
\begin{bmatrix}
    Z\\
    Y
\end{bmatrix}
\sim {\rm N} \left(
\begin{bmatrix}
    0\\
    0
\end{bmatrix},
\begin{bmatrix}
    I_d & W^\top  \\
    W  & W W^\top + \sigma^2 I_D
\end{bmatrix} \right).
\end{eqnarray}
Denote 
\begin{eqnarray}
S = 
\begin{bmatrix}
    S_{ZZ} & S_{ZY} \\
    S_{YZ} &S_{YY} 
    \end{bmatrix} =
    \begin{bmatrix}
 \E[Z  Z^\top] &  \E[Z Y^\top ]\\
   \E[Y Z^\top]  & \E[Y Y^\top]
    \end{bmatrix} \nonumber \\= 
\begin{bmatrix}
    I_d & W^\top \\
    W & W W^\top + \sigma^2 I_D
\end{bmatrix}.
\end{eqnarray}
The posterior distribution $p(Z|Y, W)$ can be obtained by linear regression of $Z$ on $Y$, $[Z|Y, W] \sim {\rm N}(\beta Y, V)$, where 
\begin{eqnarray} 
&&\beta = S_{ZY} S_{YY}^{-1}, \\
&& V = S_{ZZ} - S_{ZY} S_{YY}^{-1} S_{YZ}.
\end{eqnarray}
The above computation can be carried out by the sweep operator on $S$, with $S_{YY}$ being the pivotal matrix. 

Suppose we have observations $\{Y_i, i = 1, ..., n\}$. In the E-step, we compute 
\begin{eqnarray}
&&\E[Z_i|Y_i, W] = \beta Y_i, \label{eq:E1}\\
&&\E[Z_i Z_i^\top|Y_i, W] = V + \beta Y_i  Y_i^\top \beta^\top \label{eq:E2}.
\end{eqnarray} 
In the M-step, we compute 
\begin{eqnarray}
&&\S =
\begin{bmatrix}
    \S_{ZZ} & \S_{ZY} \\
   \S_{YZ} &\S_{YY} 
    \end{bmatrix} \nonumber  \\&=&
\begin{bmatrix}
    \sum_{i=1}^{n} \E[Z_i Z_i^\top] /n & \sum_{i=1}^{n} \E[Z_i] Y_i^\top /n\\
    \sum_{i=1}^{n} Y_i \E[Z_i]^\top/n  & \sum_{i=1}^{n} Y_i Y_i^\top/n
    \end{bmatrix}, 
\end{eqnarray}
where we use $\E[Z_i]$ and $\E[Z_i Z_i^\top]$ to denote the conditional expectations in (\ref{eq:E1}) and (\ref{eq:E2}). 
Then we regress $Y$ on $Z$ to obtain the coefficient vector and residual variance-covariance matrix
\begin{eqnarray}
&&W = \S_{YZ} \S_{ZZ}^{-1} \\
&& \Sigma = \S_{YY} - \S_{YZ} \S_{ZZ}^{-1} \S_{ZY}. 
\end{eqnarray}
 If $\sigma^2$ is unknown, it can be obtained by averaging the diagonal elements of $\Sigma$. The computation can again be done by the sweep operator on $\S$, with $\S_{ZZ}$ being the pivotal matrix. 

The E-step is based on the multivariate linear regression of $Z$ on $Y$ given $W$. The M-step updates $W$ by the multivariate linear regression of $Y$ on $Z$. Both steps can be accomplished by the sweep operator. We use the notation $S$ and $\S$ for the Gram matrices to highlight the analogy between the two steps. The EM algorithm can then be considered alternating linear regression or alternating sweep operation, which serves as a prototype for alternating back-propagation.

\bigskip
\bibliographystyle{aaai}
\bibliography{mybibfile}

\begin{thebibliography}{}

\bibitem[\protect\citeauthoryear{Cand{\`e}s, Romberg, and
  Tao}{2006}]{candes2006robust}
Cand{\`e}s, E.~J.; Romberg, J.; and Tao, T.
\newblock 2006.
\newblock Robust uncertainty principles: Exact signal reconstruction from
  highly incomplete frequency information.
\newblock {\em IEEE Transactions on information theory} 52(2):489--509.

\bibitem[\protect\citeauthoryear{Dempster, Laird, and
  Rubin}{1977}]{dempster1977maximum}
Dempster, A.~P.; Laird, N.~M.; and Rubin, D.~B.
\newblock 1977.
\newblock Maximum likelihood from incomplete data via the em algorithm.
\newblock {\em Journal of the Royal Statistical Society: B}  1--38.

\bibitem[\protect\citeauthoryear{Denton \bgroup et al\mbox.\egroup
  }{2015}]{denton2015deep}
Denton, E.~L.; Chintala, S.; Fergus, R.; et~al.
\newblock 2015.
\newblock Deep generative image models using a laplacian pyramid of adversarial
  networks.
\newblock In {\em NIPS},  1486--1494.

\bibitem[\protect\citeauthoryear{Dinh, Sohl-Dickstein, and
  Bengio}{2016}]{Dinh2016DensityEU}
Dinh, L.; Sohl-Dickstein, J.; and Bengio, S.
\newblock 2016.
\newblock Density estimation using real nvp.
\newblock {\em CoRR} abs/1605.08803.

\bibitem[\protect\citeauthoryear{Doretto \bgroup et al\mbox.\egroup
  }{2003}]{dorettoCWS03}
Doretto, G.; Chiuso, A.; Wu, Y.; and Soatto, S.
\newblock 2003.
\newblock Dynamic textures.
\newblock {\em IJCV} 51(2):91--109.

\bibitem[\protect\citeauthoryear{{Dosovitskiy}, {Springenberg}, and
  {Brox}}{2015}]{Alexey2015}
{Dosovitskiy}, E.; {Springenberg}, J.~T.; and {Brox}, T.
\newblock 2015.
\newblock Learning to generate chairs with convolutional neural networks.
\newblock In {\em CVPR}.

\bibitem[\protect\citeauthoryear{Girolami and
  Calderhead}{2011}]{girolami2011riemann}
Girolami, M., and Calderhead, B.
\newblock 2011.
\newblock Riemann manifold langevin and hamiltonian monte carlo methods.
\newblock {\em Journal of the Royal Statistical Society: B} 73(2):123--214.

\bibitem[\protect\citeauthoryear{Goodfellow \bgroup et al\mbox.\egroup
  }{2014}]{goodfellow2014generative}
Goodfellow, I.; Pouget-Abadie, J.; Mirza, M.; Xu, B.; Warde-Farley, D.; Ozair,
  S.; Courville, A.; and Bengio, Y.
\newblock 2014.
\newblock Generative adversarial nets.
\newblock In {\em NIPS},  2672--2680.

\bibitem[\protect\citeauthoryear{Hyv{\"a}rinen, Karhunen, and
  Oja}{2004}]{hyvarinen2004independent}
Hyv{\"a}rinen, A.; Karhunen, J.; and Oja, E.
\newblock 2004.
\newblock {\em Independent component analysis}.
\newblock John Wiley \& Sons.

\bibitem[\protect\citeauthoryear{Ioffe and Szegedy}{2015}]{Ioffe2015BatchNA}
Ioffe, S., and Szegedy, C.
\newblock 2015.
\newblock Batch normalization: Accelerating deep network training by reducing
  internal covariate shift.
\newblock In {\em ICML}.

\bibitem[\protect\citeauthoryear{Kim and Park}{2008}]{kim2008nonnegative}
Kim, H., and Park, H.
\newblock 2008.
\newblock Nonnegative matrix factorization based on alternating nonnegativity
  constrained least squares and active set method.
\newblock {\em SIAM Journal on Matrix Analysis and Applications}
  30(2):713--730.

\bibitem[\protect\citeauthoryear{Kingma and Welling}{2014}]{KingmaCoRR13}
Kingma, D.~P., and Welling, M.
\newblock 2014.
\newblock Auto-encoding variational bayes.
\newblock In {\em ICLR}.

\bibitem[\protect\citeauthoryear{Koren, Bell, and
  Volinsky}{2009}]{koren2009matrix}
Koren, Y.; Bell, R.; and Volinsky, C.
\newblock 2009.
\newblock Matrix factorization techniques for recommender systems.
\newblock {\em Computer} 42(8):30--37.

\bibitem[\protect\citeauthoryear{Krizhevsky, Sutskever, and
  Hinton}{2012}]{krizhevsky2012imagenet}
Krizhevsky, A.; Sutskever, I.; and Hinton, G.~E.
\newblock 2012.
\newblock Imagenet classification with deep convolutional neural networks.
\newblock In {\em NIPS},  1097--1105.

\bibitem[\protect\citeauthoryear{LeCun \bgroup et al\mbox.\egroup
  }{1998}]{lecun1998gradient}
LeCun, Y.; Bottou, L.; Bengio, Y.; and Haffner, P.
\newblock 1998.
\newblock Gradient-based learning applied to document recognition.
\newblock {\em Proceedings of the IEEE} 86(11):2278--2324.

\bibitem[\protect\citeauthoryear{Lee and Seung}{2001}]{lee2001algorithms}
Lee, D.~D., and Seung, H.~S.
\newblock 2001.
\newblock Algorithms for non-negative matrix factorization.
\newblock In {\em NIPS},  556--562.

\bibitem[\protect\citeauthoryear{Liu \bgroup et al\mbox.\egroup
  }{2015}]{liu2015deep}
Liu, Z.; Luo, P.; Wang, X.; and Tang, X.
\newblock 2015.
\newblock Deep learning face attributes in the wild.
\newblock In {\em ICCV},  3730--3738.

\bibitem[\protect\citeauthoryear{Liu, Rubin, and Wu}{1998}]{liu1998parameter}
Liu, C.; Rubin, D.~B.; and Wu, Y.~N.
\newblock 1998.
\newblock Parameter expansion to accelerate em: The px-em algorithm.
\newblock {\em Biometrika} 85(4):755--770.

\bibitem[\protect\citeauthoryear{Lu, Zhu, and Wu}{2016}]{LuZhuWu2016}
Lu, Y.; Zhu, S.-C.; and Wu, Y.~N.
\newblock 2016.
\newblock Learning {FRAME} models using {CNN} filters.
\newblock In {\em AAAI}.

\bibitem[\protect\citeauthoryear{Maas, Hannun, and
  Ng}{2013}]{maas2013rectifier}
Maas, A.~L.; Hannun, A.~Y.; and Ng, A.~Y.
\newblock 2013.
\newblock Rectifier nonlinearities improve neural network acoustic models.
\newblock In {\em ICML}.

\bibitem[\protect\citeauthoryear{McDermott and
  Simoncelli}{2011}]{mcdermott2011sound}
McDermott, J.~H., and Simoncelli, E.~P.
\newblock 2011.
\newblock Sound texture perception via statistics of the auditory periphery:
  evidence from sound synthesis.
\newblock {\em Neuron} 71(5):926--940.

\bibitem[\protect\citeauthoryear{Mnih and Gregor}{2014}]{MnihGregor2014}
Mnih, A., and Gregor, K.
\newblock 2014.
\newblock Neural variational inference and learning in belief networks.
\newblock In {\em ICML}.

\bibitem[\protect\citeauthoryear{Neal}{2011}]{neal2011mcmc}
Neal, R.~M.
\newblock 2011.
\newblock Mcmc using hamiltonian dynamics.
\newblock {\em Handbook of Markov Chain Monte Carlo} 2.

\bibitem[\protect\citeauthoryear{Olshausen and
  Field}{1997}]{olshausen1997sparse}
Olshausen, B.~A., and Field, D.~J.
\newblock 1997.
\newblock Sparse coding with an overcomplete basis set: A strategy employed by
  v1?
\newblock {\em Vision Research} 37(23):3311--3325.

\bibitem[\protect\citeauthoryear{Pascanu, Montufar, and
  Bengio}{2013}]{pascanu2013number}
Pascanu, R.; Montufar, G.; and Bengio, Y.
\newblock 2013.
\newblock On the number of response regions of deep feed forward networks with
  piece-wise linear activations.
\newblock {\em arXiv:1312.6098}.

\bibitem[\protect\citeauthoryear{Radford, Metz, and
  Chintala}{2016}]{radford2015unsupervised}
Radford, A.; Metz, L.; and Chintala, S.
\newblock 2016.
\newblock Unsupervised representation learning with deep convolutional
  generative adversarial networks.
\newblock In {\em ICLR}.

\bibitem[\protect\citeauthoryear{Rezende, Mohamed, and
  Wierstra}{2014}]{RezendeICML2014}
Rezende, D.~J.; Mohamed, S.; and Wierstra, D.
\newblock 2014.
\newblock Stochastic backpropagation and approximate inference in deep
  generative models.
\newblock In {\em NIPS},  1278--1286.

\bibitem[\protect\citeauthoryear{Roweis and Saul}{2000}]{roweis2000nonlinear}
Roweis, S.~T., and Saul, L.~K.
\newblock 2000.
\newblock Nonlinear dimensionality reduction by locally linear embedding.
\newblock {\em Science} 290(5500):2323--2326.

\bibitem[\protect\citeauthoryear{Rubin and Thayer}{1982}]{rubin1982algorithms}
Rubin, D.~B., and Thayer, D.~T.
\newblock 1982.
\newblock Em algorithms for ml factor analysis.
\newblock {\em Psychometrika} 47(1):69--76.

\bibitem[\protect\citeauthoryear{Rubin}{2004}]{rubin2004multiple}
Rubin, D.~B.
\newblock 2004.
\newblock {\em Multiple imputation for nonresponse in surveys}, volume~81.
\newblock John Wiley \& Sons.

\bibitem[\protect\citeauthoryear{Vedaldi and Lenc}{2015}]{matconvnn}
Vedaldi, A., and Lenc, K.
\newblock 2015.
\newblock Matconvnet -- convolutional neural networks for matlab.
\newblock In {\em Int. Conf. on Multimedia}.

\bibitem[\protect\citeauthoryear{Vincent \bgroup et al\mbox.\egroup
  }{2008}]{vincent2008extracting}
Vincent, P.; Larochelle, H.; Bengio, Y.; and Manzagol, P.-A.
\newblock 2008.
\newblock Extracting and composing robust features with denoising autoencoders.
\newblock In {\em ICML},  1096--1103.

\bibitem[\protect\citeauthoryear{Wang and Zhu}{2003}]{Wang03modelingtextured}
Wang, Y., and Zhu, S.-C.
\newblock 2003.
\newblock Modeling textured motion: Particle, wave and sketch.
\newblock In {\em ICCV},  213--220.

\bibitem[\protect\citeauthoryear{Xie \bgroup et al\mbox.\egroup
  }{2016}]{XieLuICML}
Xie, J.; Lu, Y.; Zhu, S.-C.; and Wu, Y.~N.
\newblock 2016.
\newblock A theory of generative convnet.
\newblock In {\em ICML}.

\bibitem[\protect\citeauthoryear{Xu \bgroup et al\mbox.\egroup
  }{2015}]{Xu2015EmpiricalEO}
Xu, B.; Wang, N.; Chen, T.; and Li, M.
\newblock 2015.
\newblock Empirical evaluation of rectified activations in convolutional
  network.
\newblock {\em CoRR} abs/1505.00853.

\bibitem[\protect\citeauthoryear{Younes}{1999}]{younes1999convergence}
Younes, L.
\newblock 1999.
\newblock On the convergence of markovian stochastic algorithms with rapidly
  decreasing ergodicity rates.
\newblock {\em Stochastics: An International Journal of Probability and
  Stochastic Processes} 65(3-4):177--228.

\end{thebibliography}

\end{document}